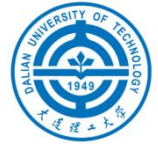

# Who will dropout from university? Academic risk prediction based on interpretable machine learning

Shudong YANG[1]

(1. Dalian University of Technology, Dalian, China, 116024)

**Abstract**: In the institutional research mode, in order to explore which characteristics are the best indicators for predicting academic risk from the student behavior data sets that have high-dimensional, unbalanced classified small sample, it transforms the academic risk prediction of college students into a binary classification task. It predicts academic risk based on the LightGBM model and the interpretable machine learning method of Shapley value. The simulation results show that from the global perspective of the prediction model, characteristics such as the quality of academic partners, the seating position in classroom, the dormitory study atmosphere, the English scores of the college entrance examination, the quantity of academic partners, the addiction level of video games, the mobility of academic partners, and the degree of truancy are the best 8 predictors for academic risk. It is contrary to intuition that characteristics such as living in campus or not, work-study, lipstick addiction, student leader or not, lover amount, and smoking have little correlation with university academic risk in this experiment. From the local perspective of the sample, the factors affecting academic risk vary from person to person. It can perform personalized interpretable analysis through Shapley values, which cannot be done by traditional mathematical statistical prediction models. The academic contributions of this research are mainly in two aspects: First, the learning interaction networks is proposed for the first time, so that social behavior can be used to compensate for the one-sided individual behavior and improve the performance of academic risk prediction. Second, the introduction of Shapley value calculation makes machine learning that lacks a clear reasoning process visualized, and provides intuitive decision support for education managers.

# 1. Introduction

It poses new challenges to the quality of student management with the expansion of undergraduate enrollment in China. Discovering academic risks in time through technical means, and preventing them before they happen, has certain practical and social significance for student management business.

Academic risk refers to the risk of dropping out, delaying or failing to obtain a degree without completing university studies (Huang Zhaoxu, etc., 2011)[1]. There are 4 risk control methods in management include risk averse, loss control, risk transfer, and risk self-retention (Yan Chunning, 2002)[2]. However, these methods are not fully applicable to the field of higher education management: Generally, the academic risks of college students cannot be avoided completely, and the academic risks cannot be transferred or hedged through contracts or insurance (Lai Desheng, 2009)[3]. So it can only do risk averse before the event, loss control during the event, and risk self-retention after the event. The scope of this research will focus on the use of data technology to predict the academic risks of undergraduates, so as to meet the actual business needs of university counselors for the timeliness, accuracy, model interpretability, and sample interpretability of the forecast. In the institutional research mode, relevant data generally have the characteristics of high dimensions, small samples (Mcneish, 2017)[4], and imbalanced classification (Jeon, 2009)[5]. The research question of this article is, which characteristics are the best indicators for predicting academic risk from the student behavior data sets that have high-dimensional, unbalanced classified small sample?

## 2. Related work

### 2.1 Data features

From the business perspective of university counselors, in the era of mobile Internet, undergraduate student management is generally based on grid management with the boundary of the grade group (the same grade and the same major) (Hong Lei et al., 2017)[6]. Therefore, the academic risk data set generated under this management paradigm generally has the following characteristics:

- **Small sample with heterogeneity**. The size of the grade group generally does not exceed 200, and students of different majors have certain characteristics differences, and the non-adjacent grades have certain inter-generational differences, and it is impossible to generalize the principle of universality.
- **High-dimensional data**. Variables related to academic performance can range from dozens to hundreds of thousands. Some data is difficult to automatically collect, such as love times; for unstructured data, characteristics engineering are needed; These high-dimensional variables may have correlations.
- **Unbalanced data**. Generally, the proportion of unusual changes such as academic risks is relatively small (Ma Xin et al., 2012)[7]. It needs special treatment before data analysis and performance evaluation..

From the perspective of data flow, there is the possibility of missing or even wrong collection due to certain academic behaviors are concealed in the data collection stage(Qu Shuaifeng, 2020)[8]. In the data preprocessing stage and data analysis stage, behavior calculations have a certain degree of uncertainty due to the existence of upstream data collection deviations and inherent deviations of feature engineering. Finally, in the data application stage, the processed data, that is, the results of academic risk prediction return to the students as the data source, forming a closed data loop. The longer the period of learning input and output, the stronger the lag of the prediction. Therefore, academic risk prediction is a long-term complex prediction, and it is difficult to accurately predict its final development outcome, but it can give a predictable range in a specific time period (Kelly, 1994)[9].

---

[1] Huang Zhaoxu, Zheng Xiaoqi. Research on the Risk of Personal Higher Education Investment[J].Higher Education Development and Evaluation, 2011, 27(04):34-37+100+122.
[2] Yan Chunning. Risk Management[M]. Shanghai University Press, 2002.
[3] Lay Desheng. Higher Education Investment: Risk and Prevention[J]. Journal of Beijing Normal University (Social Science), 2009(2):86-91.
[4] Mcneish D . Challenging Conventional Wisdom for Multivariate Statistical Models With Small Samples[J]. Review of Educational Research, 2017, 87(6).
[5] Jeon M J , Lee G , Hwang J W , et al. Estimating reliability of school-level scores using multilevel and generalizability theory models[J]. Asia Pacific Education Review, 2009, 10(2):149-158.
[6] Hong Lei, Zhang Pei. Constructing the Grid Management Mode of College Students under Massive Data Background[J]. Modern Education Management, 2017(12):96-101.
[7] Ma Xin, Wang Xue, Yang Yang. Prediction of degradation for undergraduate using random forest[J]. Journal of Jiangsu University of Science and Technology,2012,26(1):86-90. DOI:10.3969/j.issn.1673-4807.2012.01.018
[8] Qu Shuaifeng. Strategies for Helping Students with Learning Difficulties in Colleges and Universities Based on the Life Cycle Theory[J]. Journal of Wuhan University of Technology (Social Sciences Edition), 2020,33(3):130-136.

## 2.2 Theoretical basis of data analysis

Educational economics believes that the process and results obtained by college students in receiving higher education are input and output respectively (Fan Xianzuo, 1999)[10]. It is an extension of the application of input-output theory from economics in the field of higher education. It has produced derivative concepts such as "learning input", "learning output", and "academic risk". The input-output theory provides a theoretical basis for explaining the relationship between the three.

The initial research focus of learning investment was behavior investment and learning time investment, and later psychological investment such as emotional investment and cognitive investment was introduced (Ainley, 1993)[11]. Nowadays, the academic circles generally believe that the essence of learning input is the psychological input to understand and master the learning content, and the learning behavior input is the external manifestation of psychological input (Zou Min et al., 2013)[12]. Learning output is the development status and results of the knowledge, ability and other aspects acquired by students after participating in learning activities in some form (Eisner, 1979)[13]. The main short-term external performance of learning output is academic performance. Learning input and learning output are non-linear, and there is a diminishing marginal return effect (Cui Weiguo, 2000)[14]. Academic risk is a potential negative learning output, which is caused by uncertainty or insufficient learning input (Long Qi et al., 2020)[15].

## 2.3 Methodology of data analysis

Traditional mathematical statistics methods based on probability theory have rich application practices in processing small sample data sets. For the regression problem of academic risk prediction, the linear regression model is mostly used. Since its weight is set on the overall model rather than the sample, it is suitable for predicting the average academic performance as a whole (Huang et al., 2013)[16], but not suitable for For the individual's academic risk prediction. On the other hand, logistic regression models (Peng et al., 2002)[17] (Wang Liang, 2015)[18] are often used for academic risk prediction classification problems, but they are prone to under-fitting, and they have poor classification performance when the data features are missing or the feature space is large. The computing load will show a non-linear increase when the data features are too many(Wu Enda, 2017)[19], which is not suitable for the high-dimensional data set in this research scenario. The above parameter method can be used when the form of the probability distribution is known, but the sample data in practice is often not enough to support the judgment of its distribution. In this case, non-parametric methods are generally used (Alpaydin, 2018)[20].

---

[9] Kelly K . Out of control: The new biology of machines, social systems, and the economic world, Chapter 22[M]. Addison-Wesley, 1994.
[10] Fan Xianzuo. Educational Economics[M]. Beijing: People's Education Press, 1999: 62-63
[11] Ainley Mary D. Styles of engagement with learning - multidimensional assessment of their relationship with strategy use and school-achievement[J]. Journal of Educational Psychology, 1993, 85(3), 395-405.
[12] Zou Min, Tan Dingliang. A review of the research on student learning input and output evaluation[J]. Educational Measurement and Evaluation (Theoretical Edition), 2013(3):10-14. DOI:10.3969/j.issn.1674-1536.2013.03.003.
[13] Eisner E W. The educational imagination：On the design and evaluation of school programs[M]. New York:Macmillan, 1979: 125．
[14] Cui Weiguo. Economic Analysis of Learning Input and Output[J]. Journal of Beijing Institute of Technology (Social Science Edition), 2000(04):76-78.
[15] LONG Qi, NI Juan. A Study on Key Factors of Promoting College Student Engagement[J]. Journal of Educational Studies, 2020, 16(6):117-127. DOI:10.14082/j.cnki.1673-1298.2020.06.013.
[16] Huang S , Ning F . Predicting student academic performance in an engineering dynamics course: A comparison of four types of predictive mathematical models[J]. Computers & Education, 2013, 61(2):133-145.

Nearest neighbor classifiers (KNN) and classic support vector machines (SVM) are all non-parametric methods. These linear classifiers also have certain application practices in academic risk prediction (Li Haojun, 2014)[21] (Zhang Yuteng, 2018)[22]. Although these linear classifiers are suitable for small sample data sets and have good generalization performance and robustness, they cannot solve the problem of sample linear inseparability.

In order to solve the problem of linear inseparability of samples, nonlinear classifiers or deep learning methods can be used. Practice has proved that convolutional neural networks and long and short term memory recurrent neural networks (LSTM) can improve various performance indicators of academic performance prediction (Ramanathan et al., 2021)[23]. For academic data sets containing time series information, the use of bidirectional long and short term memory recurrent neural network (Bi-LSTM) can significantly improve prediction performance (Li et al, 2020)[24]. However, these deep neural networks contain many hidden layers and nonlinear operations, which cannot explain the internal operating mechanism, and are difficult to apply to individual student education decisions.

The above are all supervised learning, that is, each training sample needs to be labeled. In practice, semi-supervised machine learning methods are also used to predict academic risks. A small amount of academic risk-labeled data and a large amount of unlabeled data are used to train prediction models to improve prediction accuracy (Livieris et al, 2017)[25]. DBN deep belief network is an attempt to optimize academic performance classification performance (Sokkhey et al., 2020)[26]. However, this type of method requires massive data and is not suitable for the small sample data set in this research scenario.

With the promotion of educational data collection infrastructure, educational big data has formed a primitive accumulation. The education field has jumped from a poor data mine to a gold mine of big data (Fischer et al, 2020)[27]. In recent years, the research on interpretable machine learning methods in the field of education has gradually increased. There are two main interpretable paths. First, the prediction model itself is explained based on the overall sample, such as the visualization of the dropout prediction decision tree model to explain the internal operating mechanism of machine learning (Rodríguez-Muiz et al, 2019)[28]; Second, interpret the prediction results based on the sample individuals, such as using Shapley value to personalize the interpretation of the academic risk prediction results generated by the Catboost machine learning model, which is used for the early warning of the academic risk of individual students (Zhai et al., 2021)[29].

---

[17] Chao-Ying, Joanne, Peng, et al. The Use and Interpretation of Logistic Regression in Higher Education Journals: 1988–1999[J]. Research in Higher Education, 2002.
[18] Wang Liang. Learning Analytics: Preliminary Study of Creating Course Predictive Model[J]. Research and Exploration in Laboratory, 2015, 34(1): 215-218,246. DOI:10.3969/j.issn.1006-7167.2015.01.052.
[19] Andrew Ng. Machine Learning[EB/OL]. Coursera, 2017, <https://www.coursera.org/learn/machine-learning>
[20] Ethem Alpaydin. Introduction to Machine Learning (Original Book 3rd Edition)[M]. Machinery Industry Press, 2018:107
[21] Li Haojun, Xiang Jing, Hua Yanyan. mCSCL Learning Partner Grouping Strategy Research Based on KNN Clustering Algorithm[J]. Modern Educational Technology, 2014, 24(3): 86-93. DOI:10.3969/j.issn.1009-8097.2014.03.012.
[22] Zhang Yuteng. Recognition study on the suicidal tendency of college students based on SVM[J]. Chinese Journal of School Health, 2018,39(5): 685-687. DOI:10.16835/j.cnki.1000-9817.2018.05.013.
[23] Ramanathan K , Thangavel B . Minkowski Sommon Feature Map-based Densely Connected Deep Convolution Network with LSTM for academic performance prediction[J]. Concurrency and Computation Practice and Experience, 2021(4).
[24] Li X , Zhu X , Zhu X , et al. Student Academic Performance Prediction Using Deep Multi-source Behavior Sequential Network[C]//. Advances in Knowledge Discovery and Data Mining, 2020.
[25] Livieris I E , Drakopoulou K , Tampakas V T , et al. Predicting Secondary School Students' Performance Utilizing a Semi-supervised Learning Approach[J]. Journal of Educational Computing Research, 2018:073563311775261.
[26] Sokkhey P , Okazaki T . Development and Optimization of Deep Belief Networks Applied for Academic Performance Prediction with Larger Datasets[J]. IEIE Transactions on Smart Processing & Computing, 2020, 9.

## 2.4 Comment

At the data level, previous studies mainly used individual data as the basis for prediction. However, in a learning system, changes in the learning status of students may cause changes in the learning status of other students (Johnson, 2009)[30]. Therefore, this research explores the use of feature engineering and other methods to refine social behavior data, so that social behavior can be used to predict academic risks.

At the theoretical level, from the perspective of value-added evaluation, the current academic community does not fully consider the differences in the starting point of students. Therefore, this research adds admission scores and student characteristics on the basis of behavioral investment, learning time investment, emotional investment, and cognitive investment.

At the methodological level, for data sets with high-dimensional, unbalanced, and heterogeneous small sample characteristics, both traditional mathematical statistics and traditional machine learning methods have their drawbacks: on the one hand, traditional mathematical statistics Due to the existence of the independence assumption, it is necessary to reduce the dimensionality of the variables before modeling, and lossy compression will cause a certain loss of features; in addition, although the traditional mathematical statistics method has a certain degree of interpretability, it is limited to the model The whole, that is, the weight of a variable is for the model as a whole, not for the individual sample, so this interpretability has certain limitations. On the other hand, the general machine learning method lacks a clear reasoning process, that is, it cannot show its internal operating mechanism, and it was difficult to be introduced into educational decision-making in the past (Ding Xiaohao, 2017)[31]. In recent years, with the continuous improvement of computer computing power, GPU acceleration and the rise of neural computing, the computing power performance bottleneck caused by the use of complex machine learning models in the past is no longer obvious, such as XGBoost, GBDT, LightGBM, CatBoost and other emerging machine learning algorithms crush traditional machine learning algorithms in terms of model training and prediction performance. There is a certain negative correlation between the accuracy and interpretability of prediction models, and the growing tension between the two also promotes interpretability. Therefore, this research explores the application of explainable machine learning methods in the field of higher education management.

---

[27] Fischer C , Pardos Z A , Baker R S , et al. Mining Big Data in Education: Affordances and Challenges[J]. Review of Research in Education, 2020, 44(1).
[28] LJ Rodríguez-Muiz, Bernardo A B , Esteban M , et al. Dropout and transfer paths: What are the risky profiles when analyzing university persistence with machine learning techniques?[J]. PLOS ONE, 2019, 14.
[29] Zhai M , Wang S , Wang Y , et al. An interpretable prediction method for university student academic crisis warning[J]. Complex & Intelligent Systems, 2021.
[30] Johnson D W, Johnson R T, Holubec E J. Circles of Learning: Cooperation in the Classroom 6th edition[M]. Interaction Book Co., 2009
[31] Ding Xiaohao. The Educational Research under the Context of Big Data: Opportunities and Challenges[J]. Tsinghua Journal of Education, 2017,38(5):8-14.

In summary, by combing through relevant past domestic and foreign research, the innovation space of this research lies in: fully considering the differences of students' starting points, and extracting social behavior data through methods such as feature engineering, and exploring interpretable machine learning methods to predict academic risk in a grid management scenario with grade group as the boundary.

# 3. Research design

## 3.1 Feature engineering

This study collected the academic data of 2015~2018 undergraduates majoring in e-commerce in a private university in China. These data were collected through the teaching management system, book borrowing system, field surveys, attendance photos, group work archives, and semi-structured in-depth interviews with university counselors and student representatives, covering a total of 446 students. Multi-source data fusion of the above data from different sources based on student number, name and face, and then data cleaning, a data set containing 37 variables and 367 effective records of individual behaviors and group behaviors was obtained.

The dependent variable is a binary variable with or without academic risk. This value is taken from the student's grade point average (hereinafter referred to as GPA). The lower limit of GPA is 0 and the upper limit is 5. After the second grade, when the value is less than 2, students will receive an academic risk warning. Therefore, this study adopts the same rules The continuous quantity from the interval of [0,5] is mapped to the binary classification set of {0,1}, where "0" represents no academic risk, and "1" represents academic risk. The distribution ratio of the two is 0.845:0.155, which is Unbalanced classification.

Compared with traditional statistical methods, the regularized machine learning method does not need to do parameter hypothesis testing, nor does it need to consider the collinearity of variables, so it does not need to do principal component analysis and other dimensionality reduction processing, all 36 variables except GPA Regarded as independent variables, the independent variables are summarized in the following two tables:

**Table 1 Summary Table of Numerical Variables**

| Type | Variable | Feature description | Mean | Std | Median | Max |
|---|---|---|---|---|---|---|
| College entrance examination scores | ExmSumN | Total score | 383.1/750.0 | 74.2 | 395.0 | 557.0 |
| | ExamCnN | Chinese score | 81.7/150.0 | 18.3 | 86.0 | 117.0 |
| | ExamEnN | Foreign language score | 66.7/150.0 | 25.8 | 66.0 | 126.0 |
| | ExamMatN | Maths score | 55.4/150.0 | 32.0 | 61.0 | 120.0 |
| | ExamProN | Professional course score | 179.2/300.0 | 46.5 | 173.0 | 296.0 |
| Social network | DgrCnt | Quantity of academic partners (degree centrality) | 0.279 | 0.138 | 0.238 | 0.688 |
| | BtwnCnt | Mobility of academic partners (betweeness centrality) | 0.018 | 0.028 | 0.007 | 0.216 |
| | EgnCnt | Quality of academic partners (eigenvector centrality) | 0.074 | 0.070 | 0.055 | 0.294 |

Table 2 Summary table of categorical variables

| Type | Variable | Feature description | Type amount | Balanced | Samples |
|---|---|---|---|---|---|
| Fixed basic information | EntrnceTyp | Entrance type | 2 | Yes | Upgrade, Common |
| | Gndr | Gender | 2 | Yes | Male, Female |
| | Age | Age at entrance | 9 | Yes | 16, 17, 18, 19, 20... |
| | UrbnRrl | Urban-rural | 4 | Yes | Rural-fresh, urban-former... |
| | BrthPrvnce | Birth province | 19 | No | Liaoning, Beijing... |
| | Eth | Ethnicity | 8 | No | Han, Mongolian... |
| | Hrscpe | Horoscope | 12 | Yes | Aries, Taurus... |
| Unfixed basic information | NonRsdnt | Non-resident | 2 | No | No, Yes |
| | Gurdn | Guardian type | 4 | No | Parents, Father, Mother, Other |
| | PltclStts | Politics status | 4 | No | Masses... |
| | Dorm | Dormitory code | 103 | Yes | Dormitory pseudocodes |
| | DrmStyle | Dormitory study atmosphere | 4 | Yes | For-fun, for-study... |
| | GftedStdnt | Gifted Student | 2 | No | No, Yes |
| | Class | Class code | 14 | Yes | Class pseudocodes |
| Learning behavior | SftSp | Soft soap | 2 | No | No, Yes |
| | Truant | Truant level | 4 | Yes | Never, seldom... |
| | Seat | Seat order | 3 | Yes | Front, middle, rear |
| | Leader | Leader | 3 | No | None, Class, School |
| | Awrds | Academic awards | 2 | No | No, Yes |
| | BkBrrw | Book borrow | 4 | No | Never, seldom... |
| | WrkStdy | Work study | 2 | No | No, Yes |
| Living behavior | Lover | Lover amount | 4 | Yes | 0, 1, 2, more |
| | Hair | Hair amount | 5 | Yes | 0, 1, 2, 3, 4 |
| | Tattoo | Tattoo | 2 | No | No, Yes |
| | CmpsLn | Campus loan | 2 | No | No, Yes |
| | Lpstck | Lipstick addiction | 4 | No | Never, seldom... |
| | Smk | Smoke addiction | 2 | Yes | No, Yes |
| | Game | Game addiction | 4 | Yes | Never, seldom... |

It added the dimensions of admission scores and student characteristics based on the concept of value-added evaluation, such as the five variables of college entrance examination scores in Table 1, the 7 fixed basic information variables and the 7 unfixed basic information variables in Table 2.

In this study, not only the individual learning behaviors of students are observed, but also the academic interaction behaviors among students in learning interaction networks. From the perspective of social network analysis, the structure composed of nodes (individual students) and edges (interactive behaviors) is observed under the learning interaction networks, and then the implicit academic sentiment is explored. The figure below is a learning behavior adjacency matrix composed of 96 students from the 2016 grade group. The sub-figure (a) is the dormitory network adjacency matrix after the student relationship is stabilized in the second semester of the sophomore year, that is, the students in the same dormitory are adjacent to each other. Sub-figures (b), (c), (d), (e) are the adjacency matrices of the four learning activity team network, that is, members in the same learning team are adjacent to each other. Sub-figure (f) is a synthesis of the above-mentioned adjacency matrix and other unrepresented matrices. From sub-figure (a) to sub-figure (e) are all "0-1" matrices, and sub-figure (f) is weighted. After social network analysis and calculation, we can get the quantity index (degree centrality) of the academic partner, the mobility index of the academic partner (betweeness centrality), and the quality index of the academic partner (eigenvector centrality), which belong to the interval [0,1] continuous variables. From the perspective of individual students, degree centrality represents the quantity of their cooperative academic partners, betweeness centrality represents their degree of connection in different learning teams, and eigenvector centrality represents the quality of their cooperative academic partners.

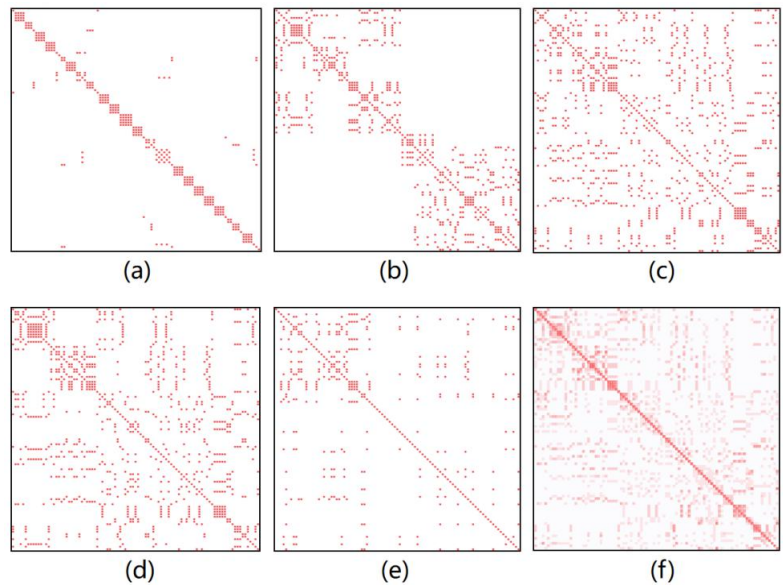

**Figure 1 The learning interaction network of 96 students in the 2016 grade group**

In order to explore which characteristics of academic behavior are the best indicators for predicting academic risk, this research exhausts as many learning and life behavior data as possible that can be collected. In order to improve the robustness and interpretability of the model, while reducing the risk of overfitting, these data are discretized. Among them, variables such as work-study programs and campus loans represent disposable income; variables such as skipping classes, seat order, and degree of borrowing represent learning attitude; variables such as the number of relationships, the degree of dependence on lipstick, smoking, and the degree of dependence on video games represent potential effects on academics performance factors.

In the later stage of model optimization, in order to improve the performance of the classification prediction model, 10 variables that have no significant impact are eliminated: Age at entrance; Birth province; Ethnicity; Horoscope; Guardian type; Dormitory code; Academic awards; Hair amount; Tattoo; Campus loan.

## 3.2 Model framework design

It performs supervised learning modeling on student datasets with academic risk labels obtained after feature engineering. The output result is the classification prediction of the student's academic risk, and then SHAP (SHapley Additive exPlanations) is used to explain the supervised learning model, and the interpretable description of the predicted result. The final outputs are the visualized SHAP value figures. The research framework and core process are shown in the figure below:

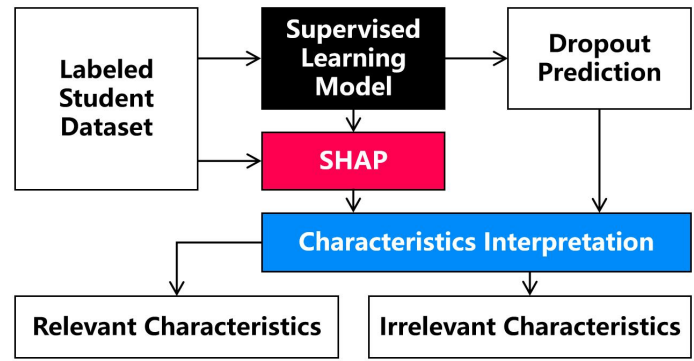

**Figure 2 Research framework and core process**

## 3.3 Selection of interpretable methods

In this study, a total of 11 common machine learning models were compared horizontally. Although the decision tree model and the K-nearest neighbor model have their own interpretability, it selected an interpretable method that is model-agnostic in order to improve the reusability. The difficulty of this research framework is the use of interpretable method that is model-agnostic to realize the decoupling of interpretable tools from the underlying machine learning model. The advantage is that it can quickly replace the machine learning model and has better scalability.

The Shapely value originated from a cooperative game: multi-person cooperation creates a certain scale of value, how to evaluate the marginal contribution of each person. This idea provides an interpretable solution for machine learning that cannot explain the operating mechanism in the past, including decision tree models and neural network models (Masís, 2021)[32]. The Shapley value is the only attribution method that satisfies the properties efficiency, symmetry, dummy and additivity, which together can be considered a definition of a fair payout (Molnar, 2020)[33].

## 3.4 Design of performance evaluation indicators

Generally speaking, when it is necessary to classify structured data, the logistic regression model can quickly output relatively reliable classification results, so it is generally used as a baseline model. It accepts the complex model if the performance of other complex models is better than the baseline model, else it rejects the complex model.

Common classification performance indicators include precision rate, recall rate, F1 value, and ROC curve. The indicators such as precision rate, recall rate, and F1 value are suitable if the dependent variable is in a balanced state, but when it is unbalanced, only the ROC curve is suitable. Generally speaking, the proportion of students who have academic risk will not be more than half, or even a very small proportion. Therefore, this study intends to adopt the ROC curve. The area under the ROC curve is the performance value, referred to as AUC (Area Under Curve).

---

[32] S Masís. Interpretable Machine Learning with Python: Learn to build interpretable high-performance models with hands-on real-world examples[M]. Packt Publishing, 2021.
[33] C Molnar. Interpretable Machine Learning: A Guide for Making Black Box Models Explainable[M]. lulu.com, 2020.

# 4. Simulation

## 4.1 Experimental result

Taking the ROC curve as the evaluation standard, it compared 11 common machine learning methods through computer simulation. The LightGBM method which achieved 0.96, 0.95, 0.95 and 0.89 on accuracy, recall, F1 value and AUC respectively, has significant advantages over logistic regression, SVM, GBDT, XGBoost, CatBoost, etc.

Sorted according to the importance of the variables, the most important predictor of academic risk is the quality of academic partners, the seating position in classroom, the dormitory study atmosphere, the English scores of the college entrance examination, the quantity of academic partners, the addiction level of video games, the mobility of academic partners, and the degree of truancy. It is contrary to intuition that characteristics such as living in campus or not, work-study, lipstick addiction, student leader or not, lover amount, and smoking have little correlation with university academic risk in this experiment.

The above performance indicators and their corresponding prediction results are undoubtedly difficult to understand for education decision makers and front-line teachers, that is, for a specific sample, humans cannot know how the characteristic value of this sample affects the final result. In recent years, explainable machine learning has gradually become an important research direction. It can prevent model bias and help decision makers understand how to use these models correctly. The more demanding scenarios, especially in the education industry, the more models are required to provide evidence to prove how they work and avoid errors, and even the specific reasons for rejecting samples need to be analyzed.

## 4.2 Interpretable analysis of predictive models

It calculated the Shapley value of the LightGBM prediction model, and obtained the Shapley value graph of the top 20 most influential variables. The higher the variable, the stronger the explanatory ability of the academic risk prediction. It can be found that the ranking of social behavior variables is higher. The quality of academic partners, the quantity of academic partners and the mobility of academic partners ranked first, sixth and eighth respectively. Each scattered point represents a sample of students, and the horizontal axis represents the degree of influence of a variable on individual academic risks. The same variable has different degrees of influence on different students, which is also in line with the actual situation. The vertical axis is meaningless, but it will jitter up and down when the Shapley value of multiple samples is repeated, which visually forms a vertical distribution. Finally, it illustrated below:

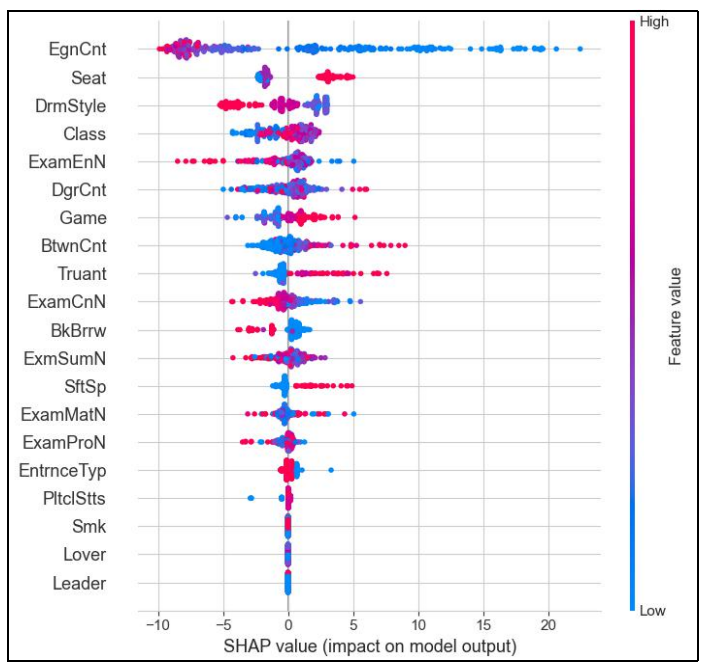

**Figure 3 Shapley value graph of the prediction model (Top 20)**

In the sample of this study, the students of this school are divided into administrative classes according to the English scores of the college entrance examination from high to low at the time of enroll, that is, the class code and the English scores of the college entrance examination are negatively correlated. From the Shapley graph, it can be observed that these two variables are closely together. It can ignore the derived variable of the class code, and only keep the English scores of the college entrance examination. The simulation results show that from the global perspective of the prediction model, characteristics such as the quality of academic partners, the seating position in classroom, the dormitory study atmosphere, the English scores of the college entrance examination, the quantity of academic partners, the addiction level of video games, the mobility of academic partners, and the degree of truancy are the best 8 predictors for academic risk. It is contrary to intuition that characteristics such as living in campus or not, work-study, lipstick addiction, student leader or not, lover amount, and smoking have little correlation with university academic risk in this experiment.

It takes the previous three variables as examples for further analysis below. Firstly, it observes the relationship between the quality of academic partner(EgnCnt) and the academic risk, as shown in the following figure:

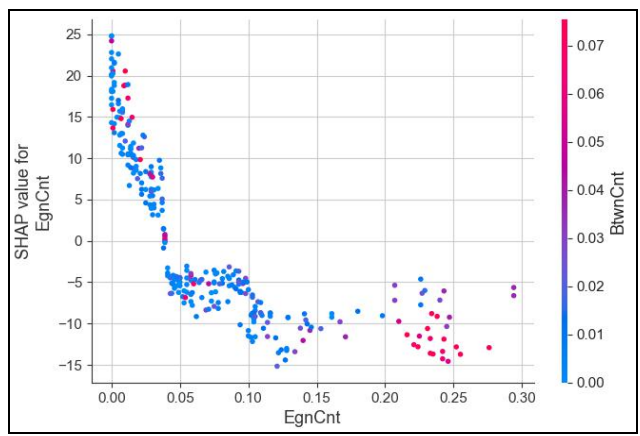

**Figure 4 Shapley dependency graph of the quality of academic partner (EgnCnt)**

From the above graph, it can be found that EgnCnt is between 0 and 0.3. The Shapley value is less than 0 when the value exceeds 0.03, which means that academic risks are not prone to occur. When it is less than 0.03, The smaller the value, the larger the Shapley value, which means that academic risks are more likely to occur. The Shapley dependency graph can also show the relationship between features. For example, students with high mobility (betweenness centrality) (pink) are mainly distributed at the poles, that is, betweenness centrality is mainly distributed in the higher and lower eigenvector centrality. A reasonable explanation for the latter is that students with low performance may often change learning groups, which form an unstable "free-riding (in Chinese)" phenomenon.

Secondly, it analyzes the relationship between the seating position in classroom and academic risk, as shown in the graph below:

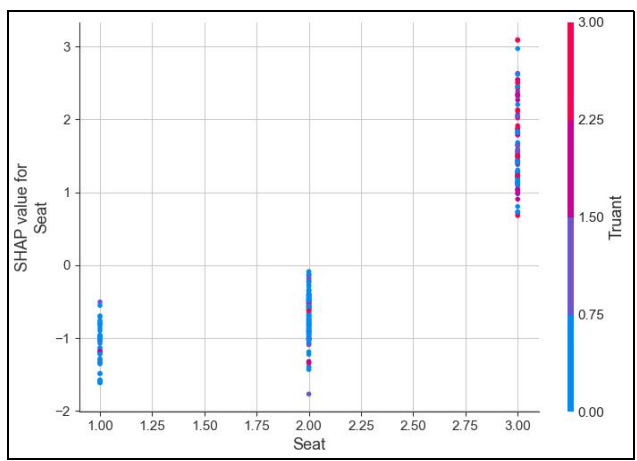

**Figure 5 Shapley dependency graph of the seating position in classroom**

From the graph above, it can be found that among the sample students' seating positions, the proportion of students sitting in the front part (encoded as 1) is small, and the proportion of students sitting in the middle part (encoded as 2) and the back part (encoded as 3) are Larger. The Shapley value of the first two is less than 0, that is, it is not easy to have academic risk, and the Shapley value of the student sitting in the back is greater than 0, that is, have academic risk. Almost all the students with a higher degree of truant (pink) are scattered in the rear seat group, indicating that the degree of truant from classes has a certain positive correlation with the seat position.

Thirdly, it analyzes the relationship between dormitory study atmosphere and academic risk, as shown in the figure below:

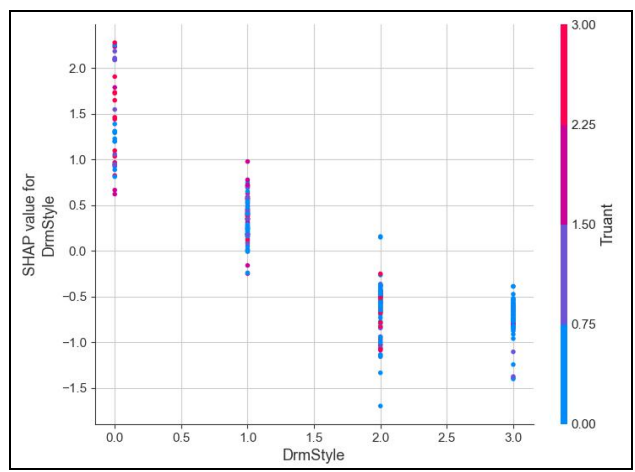

**Figure 6 Shapley dependency graph of dormitory study atmosphere**

From the above graph, it can be found that the sample dormitories with good study atmosphere (encoded as 3) and dormitories with poor study atmosphere (encoded as 0) are not as much as the other two, that is, the polarization is not obvious. Except for outliers, the Shapley values of the two types of dormitories on the right are both less than 0, which means that academic risks are not prone to occur; while the Shapley values of the two types of dormitory on the left are almost all greater than 0, which means that academic risks are prone to occur.

## 4.3 Interpretable analysis of individual case

It conducts an interpretable analysis on a certain student with academic risk. The relevant data and Shapley marginal contribution of this student are shown in the following table:

Table 3 Individual case and Shapley's marginal contribution

| Variable | Value | Interpretation | Shapley Marginal Contribution |
|---|---|---|---|
| EgnCnt | 0.002 | Poor quality of learning partners | 18.1 |
| Truant | 2 | Often skip class | 2.63 |
| Seat | 3 | Sitting at the back of the classroom | 2.51 |
| DrmStyle | 1 | Poor dormitory study atmosphere | 2.04 |
| UrbnRrl | RF | came from the countryside | -1.26 |
| ExamCnN | 0.721 | The Chinese score in college entrance examination exceeds 72.1% classmates (94) | 0.96 |
| ExamEnN | 0.733 | The English score in college entrance examination exceeds 72.1% classmates (106) | 0.95 |
| Lpstck | 3 | Always use lipstick | 0.82 |

The student's total score of the college entrance examination exceeds 82.9% classmates, but the dormitory has a poor atmosphere of study and the quality of learning partners is poor, which leads to a higher academic risk. The data visualization is shown in the figure below:

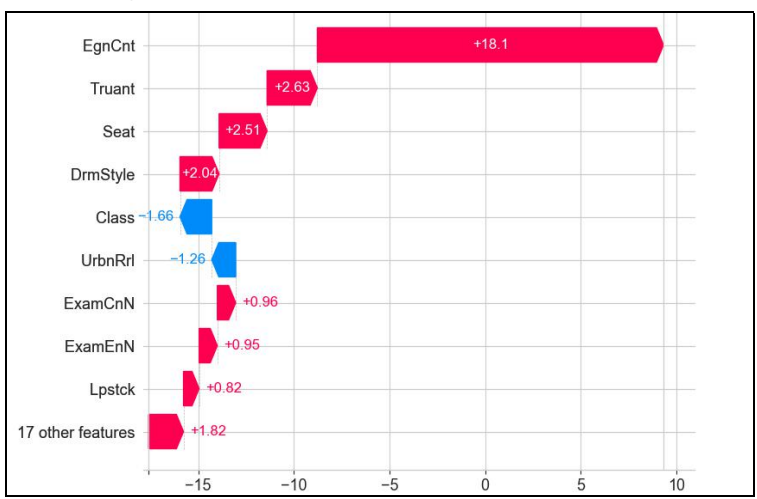

Figure 7 Shapley marginal contribution of a student's academic risk

# 5. Conclusions

This study transforms the academic risk prediction of college students into a binary classification task. The computer simulation results show that the interpretable machine learning method based on the LightGBM model and Shapley value has the best academic risk prediction performance, and it has both global interpretability (sample overall) and local interpretability (sample individual). The reasoning process of educational decision-making becomes interpretable.

The simulation results show that from the global perspective of the prediction model, characteristics such as the quality of academic partners, the seating position in classroom, the dormitory study atmosphere, the English scores of the college entrance examination, the quantity of academic partners, the addiction level of video games, the mobility of academic partners, and the degree of truancy are the best 8 predictors for academic risk. It is contrary to intuition that characteristics such as living in campus or not, work-study, lipstick addiction, student leader or not, lover amount, and smoking have little correlation with university academic risk in this experiment. From the local perspective of the sample, the factors affecting academic risk vary from person to person. It can perform personalized interpretable analysis through Shapley values, which cannot be done by traditional mathematical statistical prediction models.

The academic contributions of this research are mainly in two aspects: First, the learning interaction networks is proposed for the first time, so that social behavior can be used to compensate for the one-sided individual behavior and improve the performance of academic risk prediction. Second, the introduction of Shapley value calculation makes machine learning that lacks a clear reasoning process visualized, and provides intuitive decision support for education managers.

This research method has the following limitations and needs to be noted when porting to other similar scenarios: First, although the collinearity of input features will not affect the predictive performance of machine learning, from the perspective of the interpretability of the model, when the input features When not independent, the order of adding features will affect model interpretation (Lundberg et al., 2017)[34]; secondly , machine learning requires a lot of computing power. it can choose whether to use GPU acceleration strategy according to the complexity of the model and the size of the dataset. Finally, the lower limit of feature engineering determines the upper limit of predictive model performance.

The future research plan is to use data augmentation technologies, such as Generative Adversarial Networks (GAN), to automatically expand the data set to support the implementation conditions of deep learning methods; second, to introduce time series analysis, such as Recurrent Neural Networks (RNN) ), Long and Short-Term Memory Neural Network (LSTM), etc., used to predict academic upstarts; third, enrich the methodology and application practice of learning interaction networks.

# Acknowledgments

Thank my students and ex-colleagues for providing the raw data analyzed in this study, including but not limited to:

---

[34] Lundberg S , Lee S I . A Unified Approach to Interpreting Model Predictions[C]// Nips. 2017.

Who will dropout from university? Academic risk prediction based on interpretable machine learning
Shudong YANG, Dalian University of Technology

Qingsong YUE, Yining ZHONG, Jing QIU, Xin QI, Beier LI, Yingying PAN, Jingyun XIAO, Bohao ZHANG, Haohan ZHANG, Yang WANG et al.